\documentclass{article}
\usepackage{ijcai20}

\usepackage{times}
\usepackage{soul}
\usepackage{url}
\usepackage[hidelinks]{hyperref}
\usepackage[utf8]{inputenc}
\usepackage[small]{caption}
\usepackage{graphicx}
\usepackage{amsmath}
\usepackage{amsthm}
\usepackage{booktabs}
\usepackage{algorithm}
\usepackage{algorithmic}
\usepackage{multirow}
\usepackage{bm}
\usepackage{amssymb}

\newcommand\blfootnote[1]{%
  \begingroup
  \renewcommand\thefootnote{}\footnote{#1}%
  \addtocounter{footnote}{-1}%
  \endgroup
}

\usepackage{xcolor}

\title{Diagnosing the Environment Bias in Vision-and-Language Navigation}

\author{
Yubo Zhang\footnote{Equal Contribution}\and
Hao Tan\footnotemark[1]\And 
Mohit Bansal \\
\affiliations
UNC Chapel Hill \\
\emails
\{zhangyb, haotan, mbansal\}@cs.unc.edu
}

\begin{document}

\maketitle

\begin{abstract}
Vision-and-Language Navigation (VLN) requires an agent to follow natural-language instructions, explore the given environments, and reach the desired target locations. 
These step-by-step navigational instructions are crucial when the agent is navigating new environments about which it has no prior knowledge.
Most recent works that study VLN observe a significant performance drop when tested on unseen environments (i.e., environments not used in training), indicating that the neural agent models are highly biased towards training environments.
Although this issue is considered as one of the major challenges in VLN research, it is still under-studied and needs a clearer explanation.
In this work, we design novel diagnosis experiments via environment re-splitting and feature replacement, looking into possible reasons for this environment bias.
We observe that neither the language nor the underlying navigational graph, but the low-level visual appearance conveyed by ResNet features directly affects the agent model and contributes to this environment bias in results.
According to this observation, we explore several kinds of semantic representations that contain less low-level visual information, hence the agent learned with these features could be better generalized to unseen testing environments.
Without modifying the baseline agent model and its training method, our explored semantic features significantly decrease the performance gaps between seen and unseen on multiple datasets (i.e. R2R, R4R, and CVDN)
and achieve competitive unseen results to previous state-of-the-art models.\blfootnote{Copyright International Joint Conferences on Artificial Intelligence (IJCAI). All rights reserved.}\footnote{Code, features at \url{https://github.com/zhangybzbo/EnvBiasVLN}.}

\end{abstract}

\section{Introduction}
\begin{figure}[t]
    \centering
    \includegraphics[width=0.48\textwidth]{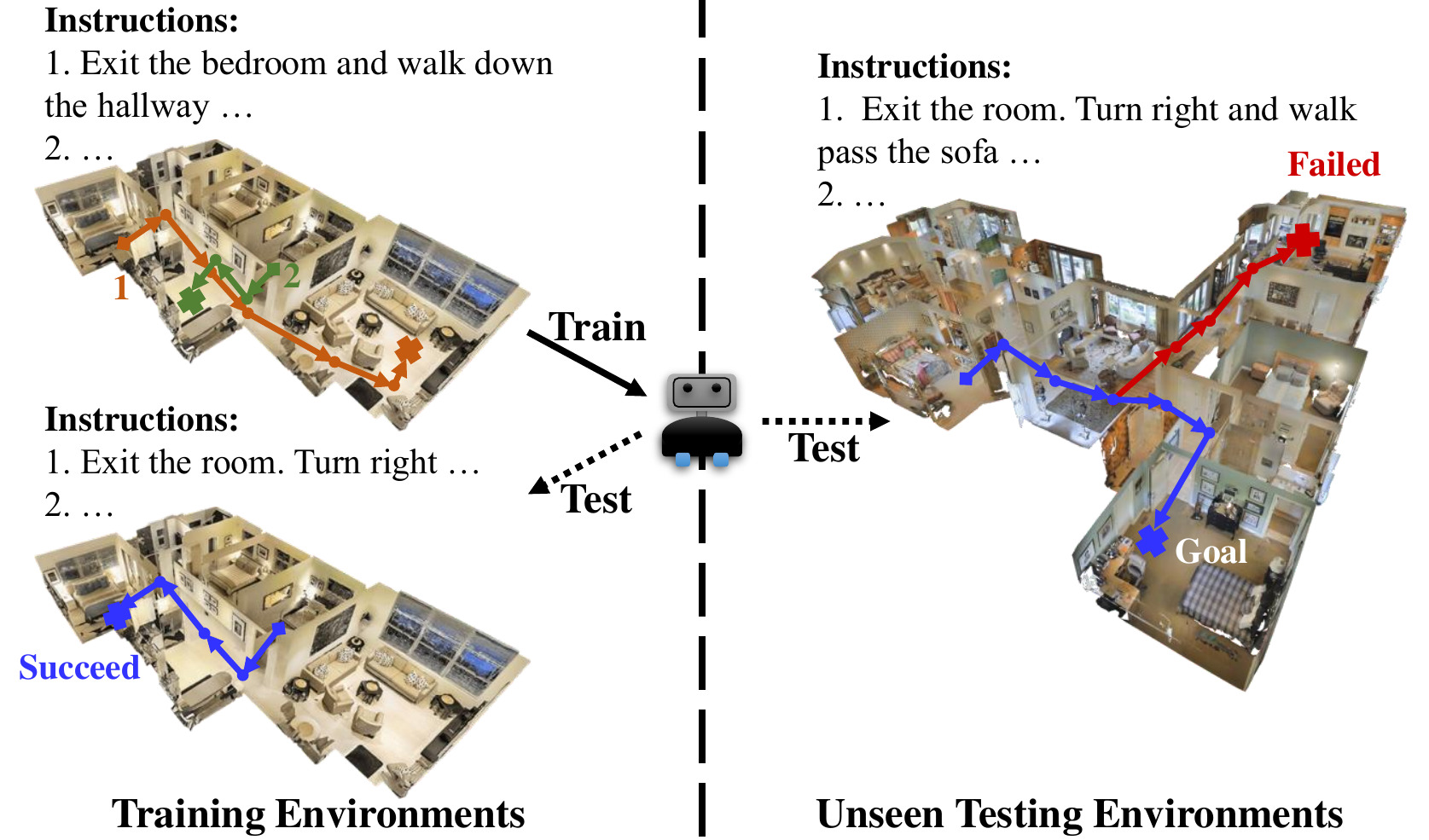}
    \caption{Vision-language-navigation: performance of the agent drops in unseen testing environments. }
    \label{fig:intro}
\end{figure}

Vision-and-Language Navigation (VLN) tests an agent's ability to understand complex natural language instructions, explore the given environments and find the correct paths to the target locations, as shown in Fig.~\ref{fig:intro}.
In this work, we focus on the instruction-guided navigation~\cite{macmahon2006walk,anderson2018vision,blukis2018mapping,chen2019touchdown} where detailed step-by-step navigational instructions are used (e.g., `Go outside the dining room and turn left ...'), in contrast to the target-oriented navigation~\cite{das2018embodied,mirowski2018learning,yu2019multi} where only the target is referred (e.g., `Go to the kitchen' or `Tell me the color of the bedroom').
Although these step-by-step instructions are over-detailed when navigating local areas (e.g., your home), they are actively used in unseen environments (e.g., a new city) where the desired target is unknown to navigational agents.
For this purpose, 
testing on unseen environments which are not used during agent-training is important and widely accepted by instruction-guided navigation datasets.

Recent works propose different methods to improve the generalizability of agents on these unseen testing environments. Most of the works \cite{anderson2018vision,wang2018look,fried2018speaker,wang2019reinforced,ma2019self,ma2019regretful,tan2019learning,huang2019transferable,hu2019you} observe a significant performance drop from the environments used in training (seen environments) to the ones not used in training (unseen environments), 
which indicates the models are strongly biased to the seen environments.
While this performance gap is emphasized as one of the major challenges,
the issue is still left unresolved and waits for an explicit explanation.
Thus, in this paper, we aim to answer three questions to this environment bias: 
1. \textbf{Where} is the bias located? 
2. \textbf{Why} does it exist? 
3. \textbf{How} to alleviate it? 

To locate \textbf{where} the bias is, 
we start by excluding natural-language instructions and underlying navigational graphs from the direct reason for this performance gap.
Then in order to conduct a detailed investigation of environments' effect on agents' performance, 
we re-split the environment and categorize the validation data into three sets based on their visibility to the training set: 
\emph{path-seen}, 
\emph{path-unseen},
and \emph{env-unseen}.
By showing that the results gradually decrease from path-seen to env-unseen data,  we thus characterize the `spatial localities' of environment bias at three levels: path level, region level, and environment level.
The \emph{low-level information} carried by the ResNet features~\cite{he2016deep} is our suspect of \textbf{why} this locality further leads to the performance gap.
By conducting the investigation of replacing the ResNet features with the $1000$ ImageNet classification probabilities, we notice that these noisy but higher-level semantic features can effectively reduce the performance gap while maintaining moderate performance on various VLN datasets (i.e., Room-to-Room, R4R, and CVDN), 
which support our hypothesis.

Following the practice of using the semantic feature in previous investigation experiments,
we further provide a discussion on \textbf{how} the environment bias could be reduced.
Three kinds of more rational and advanced semantic features are adopted:
(1) areas of detected objects~\cite{ren2015faster}; (2) ground-truth semantic views~\cite{chang2017matterport3d}; and (3) learned semantic view features. 
The results show that all of these semantic features significantly reduce the environment bias in multiple datasets and also achieve strong performance in unseen environments. 
We hope this work encourages more investigation and research into improving the generalization of vision-language models to unseen real-world scenarios.

\section{Related Work}
\noindent\textbf{Vision-and-Language Navigation}: Vision-and-language navigation is an emerging area in multi-modality research. 
Several datasets have been released recently, such as Room-to-Room~\cite{anderson2018vision}, Room-for-Room~\cite{jain2019stay}, Touchdown~\cite{chen2019touchdown}, CVDN~\cite{thomason2019vision} and EQA~\cite{das2018embodied}. 
Previous works~\cite{thomason2018shifting,wang2018look,fried2018speaker,wang2019reinforced,ma2019self,ma2019regretful,tan2019learning,hu2019you,anderson2019chasing,ke2019tactical} focusing on improving the performance of navigation models, especially in unseen testing environments, have helped to increase the navigational success rate.

\noindent\textbf{Domain Adaptation and Generalization}: The task of domain adaptation is to learn the domain invariant feature with the data samples from different domains. 
Previous works using adversarial training~\cite{goodfellow2014generative,long2018conditional} or training transfer functions~\cite{chen2019joint,rozantsev2018residual} have achieved great success.
However, in applications where the samples from target domains may not be available (e.g., the testing environments in navigation should not be used in training), these methods are hard to apply.
In domain generalization~\cite{blanchard2011generalizing,muandet2013domain,carlucci2019domain}, the goal is to predict the labels in the previous unseen domain, and the testing data is unrevealed during training. 
In VLN, two previous methods, RCM~\cite{wang2019reinforced} and EnvDrop~\cite{tan2019learning}, explore the possibility of domain adaptation. 
Both works take the testing environments in training while RCM also uses testing instructions.
In this paper, we focus on the domain generalization problem in VLN, and try to find the reasons for the failures.

\section{Vision-and-Language Navigation and its Environment Bias}
\label{sec:problem}
We first introduce the task of Vision-and-Language Navigation (VLN) and briefly describe the neural agent models used in our work.
Next, according to the survey of previous works, we show that the environment bias is widely observed in current VLN research.

\subsection{Vision-and-Language Navigation}
\label{sec:VLN}

\paragraph{Tasks}
As shown in Fig.~\ref{fig:intro}, 
the agent in VLN task is trained to navigate a certain type of environments $\{\bm{\mathrm{E}}\}$
given the instruction $\bm{\mathrm{I}}$.
Each environment $\bm{\mathrm{E}}$ is an independent space, such as a house, and consists of a set of viewpoints. 
Each viewpoint is represented as a panoramic image and can be decomposed into separate views $\{o\}$ as inputs to neural agent models.
The viewpoints and their connectivity form the navigational graph.
In practice, 
starting off at a specific viewpoint and provided with an instruction, at each subsequent time step, the agent can observe the panoramic image of the viewpoint where it is located, and choose to move along an edge of the graph to the next node (i.e., viewpoint) or stop.
The performance of the agent is evaluated by whether it eventually reaches the target location indicated by the instruction.

\paragraph{Neural Agent Models}
Most instruction-guided navigational agents are built based on attentive encoder-decoder models~\cite{bahdanau2014neural}.
Based on the instructions which are encoded by the encoder, as well as perceived environments, the decoder generates action outputs.
Since the primal focus of this work is to understand the environment bias, we use a standard neural agent model that achieves comparable results to previous works.
Specifically, we adopt the panoramic-view neural agent model in \cite{fried2018speaker} (`Follower') with modifications from \cite{tan2019learning} as our baseline.
We also exclude advanced training techniques (i.e., reinforcement learning and data augmentation), only train the agent with imitation learning in all our experiments for the same purpose.
Refer to the original papers for more details.

\begin{table}[t]\small
\newcommand{\tabincell}[2]{\begin{tabular}{@{}#1@{}}#2\end{tabular}}
\begin{center}
\begin{tabular}{lccc}
\multirow{2}{*}{\bf Method} & \multicolumn{3}{c}{\bf Result} \\
\cmidrule(lr){2-4}
 & \multicolumn{1}{c}{\bf Val Seen}  &\multicolumn{1}{c}{\bf Val Unseen} & 
\multicolumn{1}{c}{\bf Gap $\lvert \Delta\rvert$ } \\ 
\midrule
\midrule 
\multicolumn{4}{c}{\tabincell{c}{Room-to-Room~\cite{anderson2018vision}}} \\
\midrule
                R2R
                                        & 38.6  & 21.8  & 16.8  \\
                RPA 
                                        & 42.9  & 24.6  & 18.3  \\
                S-Follower
                                        & 66.4  & 35.5  & 30.9  \\
                RCM
                                        & 66.7  & 42.8  & 23.9  \\
                SMNA
                                        & 67    & 45    & 22    \\
                Regretful
                                        & 69    & 50    & 19    \\
                EnvDrop
                                        & 62.1  & 52.2  & 9.9   \\
                ALTR
                                        & 55.8  & 46.1  & 9.7   \\
                RN+Obj
                                        & 59.2  & 39.5  & 19.7   \\
                CG
                                        & 31    & 31    & \textbf{0}   \\
                Our baseline          & 56.1  & 47.5  & 8.6    \\
                {\bf Our Learned-Seg}            & 52.6  & 53.3  & \textbf{0.7}   \\
\midrule 
\midrule 
\multicolumn{4}{c}{\tabincell{c}{Room-for-Room~\cite{jain2019stay}}} \\
\midrule
                S-Follower      & 51.9  & 23.8  & 28.1  \\
                RCM                   & 55.5  & 28.6  & 26.9  \\
                Our baseline         & 54.6  & 30.7  & 23.9    \\
                \textbf{Our Learned-Seg}                  & 38.0  & 34.3  & \textbf{3.7}      \\
\midrule
\midrule 
\multicolumn{4}{c}{\tabincell{c}{CVDN~\cite{thomason2019vision}}} \\
\midrule
                NDH                   & 5.92  & 2.10  & 3.82  \\
                Our baseline           & 6.60   & 3.05  & 3.55    \\
                {\bf Our Learned-Seg}            & 5.82  & 4.41  & \textbf{1.41}  \\
\midrule
\midrule 
\multicolumn{4}{c}{\tabincell{c}{Touchdown~\cite{chen2019touchdown}}} \\
\midrule
                GA (original)      & 7.9 (dev)   & 5.5 (test)  & --    \\
                RCONCAT (original) & 9.8 (dev)   & 10.7 (test)  & --    \\
                Our baseline (original) & 15.0 (dev)   & 14.2 (test)  & --    \\
                Our baseline (re-split) & 17.5  & 5.3   & 12.2    \\
\end{tabular}
\caption{Results show the performance gaps between seen (`Val Seen') and unseen (`Val Unseen') environments in several VLN tasks. 
Room-to-Room and Room-for-Room are evaluated with `Success Rate', CVDN is evaluated with `Goal Progress', Touchdown is evaluated with `Task Completion'.
}
\label{previous-work}
\end{center}
\end{table}

\subsection{Environment Bias in Indoor Navigation}
\label{sec-3-2-envbias}
In order to evaluate the generalizability of agent models, indoor VLN datasets (e.g., those collected from Matterport3D~\cite{chang2017matterport3d}) use disjoint sets of environments in training and testing.
Two validation splits are provided as well: validation seen (which takes the data from training environments) and validation unseen (whose data is taken from testing environments different from the training).

In the first part of Table~\ref{previous-work}, we list most of the previous works
(R2R~\cite{anderson2018vision}, RPA~\cite{wang2018look}, S-Follower~\cite{fried2018speaker}, RCM~\cite{wang2019reinforced}, SMNA~\cite{ma2019self}, Regretful~\cite{ma2019regretful}, EnvDrop~\cite{tan2019learning}, ALTR~\cite{huang2019transferable}, RN+Obj~\cite{hu2019you}, CG~\cite{anderson2019chasing})
on the Room-to-Room dataset~\cite{anderson2018vision} and their \emph{success rate} under greedy decoding (i.e., without beam-search) on validation seen and validation unseen splits.
The large absolute gaps (from $30.9\%$ to $9.7\%$) between the results of seen and unseen environments show that current agent models on R2R suffer from environment bias.\footnote{Our work's aim is to both close the seen-unseen gap while also achieving competitive unseen results. Note that~\cite{anderson2019chasing} also achieve 0\% gap but at the trade-off of low unseen results.}
This phenomenon is also revealed in two other newly-released indoor navigation datasets, Room-for-Room (R4R)~\cite{jain2019stay} and Cooperative Vision-and-Dialog Navigation (CVDN)~\cite{thomason2019vision}. 
The significant result drops from seen to unseen environments can also be observed (i.e., $26.9\%$ on R4R and $3.82$ on CVDN), as shown in the second and third parts of Table~\ref{previous-work}.
Lastly, we show the results (`Our Learned-Seg' in Table~\ref{previous-work}) when the environment bias is effectively reduced by our learned semantic-segmentation features (described in Sec.~\ref{sec-6-3-semantic-view}), 
compared to our baselines (denoted as `Our baseline') and previous works.

\subsection{Environment Bias in Outdoor Navigation}
Since the three indoor VLN datasets in previous sections are collected from the same environments, Matterport3D~\cite{chang2017matterport3d}, in order to demonstrate the generality of this phenomenon, 
we investigate an outdoor navigation task from Touchdown dataset~\cite{chen2019touchdown}.
In the original data splits of Touchdown, the environment is not specifically divided into seen and unseen.
To explore whether the same environment bias can be observed in Touchdown, we split the city environment according to latitude and create two sub-environments: `training' and `unseen'.
Our baseline neural agent model is adapted with additional convolutional layers to fit this new task.
As shown in the last part of Table~\ref{previous-work}, the big performance gap between the 'training' and the 'unseen' sub-environment (from $17.5\%$ to $5.3\%$) still occurs on our re-split data (denoted as `re-split'), indicating that environment bias is a broad issue.
At the same time, when experimenting on the original data split (denoted as `original'), our baseline model achieves state-of-the-art results on the original `dev' set and `test' set, proving the validity of our model in this dataset.

\section{Where: Effect of Different Task Components}
\label{sec-4-where}
In Sec.~\ref{sec:problem},  we showed that current vision-and-language navigation (VLN) models are biased towards training environments.
The purpose of this section is to locate the reason of this environment bias.
As one of the early-released and well-explored VLN datasets, Room-to-Room (R2R)~\cite{anderson2018vision} is used as the diagnosing dataset in experiments.
We start by showing that two possible candidates, the natural language instructions and the underlying navigational graph, do not directly contribute to the environment bias.
Then the effect of visual environments is analyzed in detail.

\begin{figure}[t]
\centering
  \includegraphics[width=0.45\textwidth]{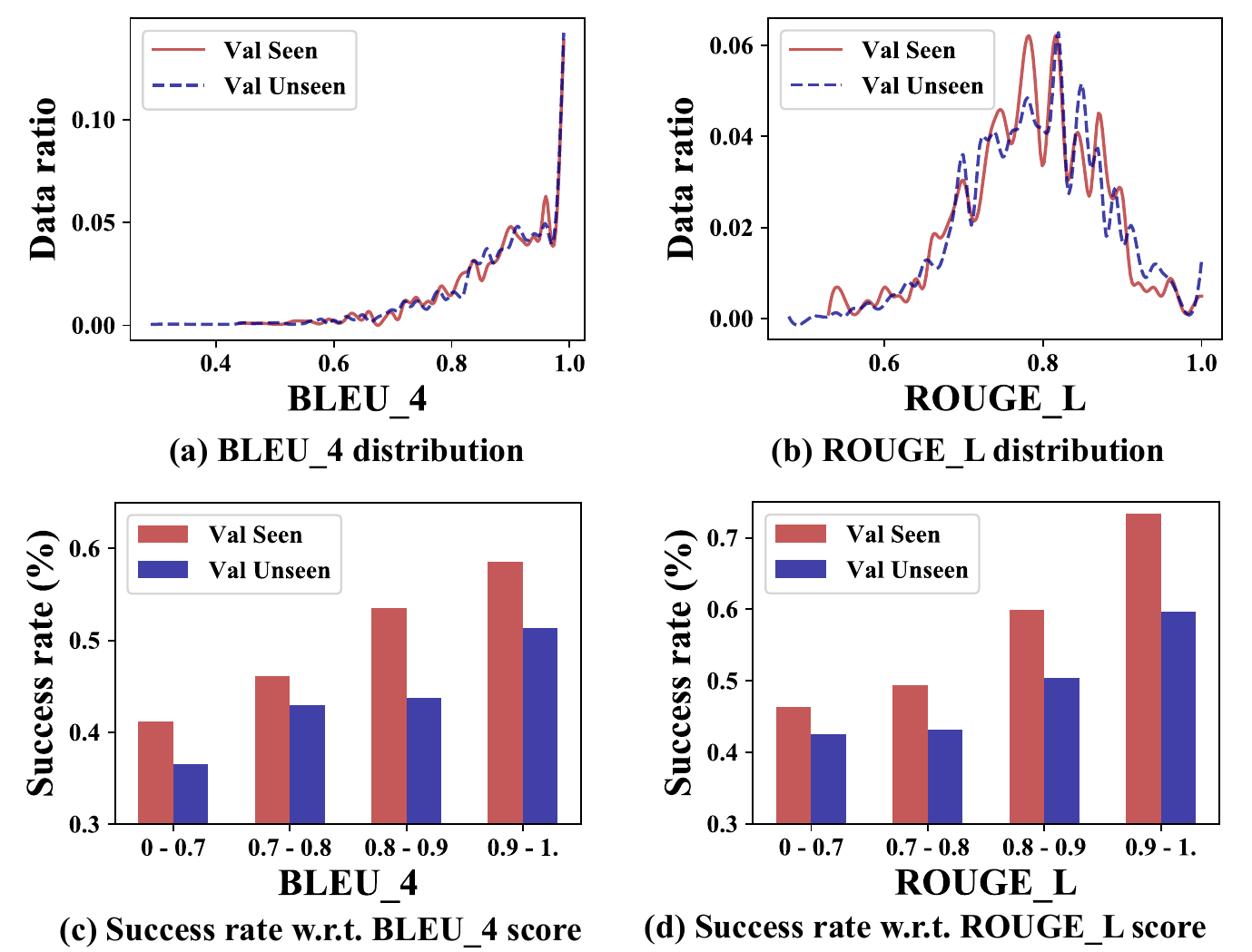} 
\caption{The language 'distance' distribution (defined by language scores) and its relationship to success rate.}
\label{Fig:lang}
\end{figure}

\subsection{The Effect of Natural-Language Instructions}
\label{sec-4-1-lang}
A common hypothesis is that 
the navigational instructions for unseen (e.g., val unseen) and training (i.e., training and val seen) environments are much different 
due to object and layout differences,
and this lingual difference can cause the performance gap.
In order to test this hypothesis,
we analyze the distributions of success rate with regard to the relationship between the instructions of training and validation sets.
To quantitatively evaluate this relationship, we define the `distances' from a validating instruction to all training instructions as the phrase-matching metric.
Suppose $x$ is a validating datum, $\mathrm{inst}(x)$ is the instruction of this datum, and $\mathbb{T}$ is the training set, we use ROUGE-L~\cite{lin2004rouge} and BLEU-4~\cite{papineni2002bleu} to calculate this `distance':
\begin{align}
\mathrm{dis}_\textsc{rouge} (x, \mathbb{T}) & = \min_{t \in \mathbb{T}}\, \mathrm{ROUGE\mbox{-}L} \left(\mathrm{inst} (x), \mathrm{inst} (t)\right) \\
\mathrm{dis}_\textsc{bleu} (x, \mathbb{T}) & =  \mathrm{BLEU\mbox{-}4} \left(\mathrm{inst} (x), \left\{\mathrm{inst} (t)\right\}_{t \in \mathbb{T}}\right)
\end{align}
where we consider all the training instructions as references in calculating the BLEU-4 score. 

The distributions of success rates and distances are shown in Fig.~\ref{Fig:lang}.
As opposed to the hypothesis, we do not observe a significant difference between the `distances' distributions  (as shown in Fig.~\ref{Fig:lang} (a, b)) of seen and unseen validation data.
As for the success rate distributions (in Fig.~\ref{Fig:lang}(c,d)), the performance is better when the instruction has smaller `distances' (i.e., higher BLEU-4/ROUGE-L scores w.r.t. the training instructions) on both validation splits.
However, comparing two splits, with the same `distance' to training instructions, seen validation still significantly outperforms the unseen validation set on success rate, 
which excludes the language from the cause of this performance gap.

\subsection{The Effect of Underlying Navigational Graph}
\begin{figure}[t]
\centering
\includegraphics[width=0.47\textwidth]{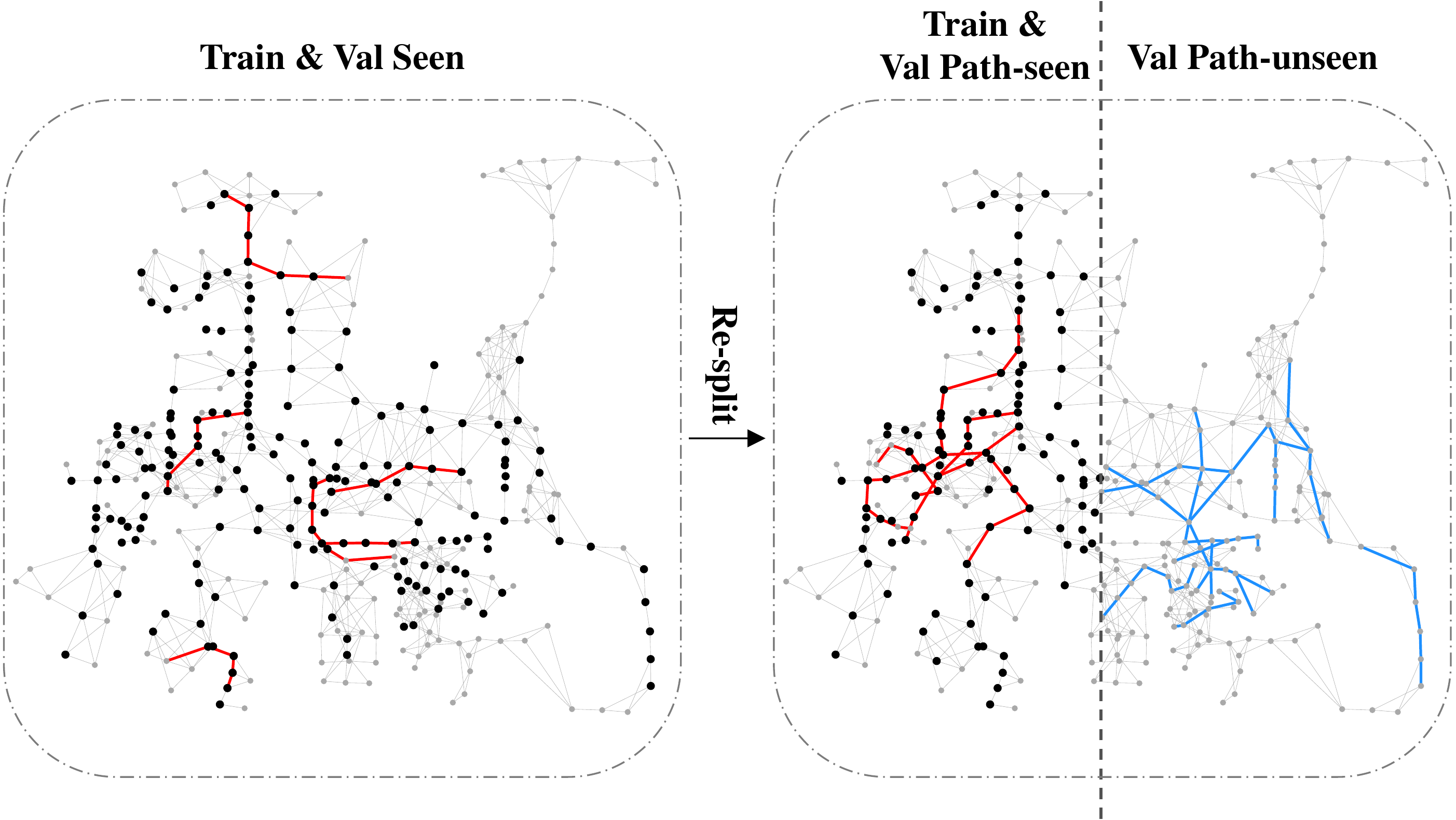} 
\caption{Graph split: left is original data and right is re-splitting data. Black vertices are viewpoints visited during training; red paths are val seen (in the left part) / val path-seen (in the right part); blue paths are val path-unseen.}
\label{Fig:graph_split}
\end{figure}

An environment could be considered as its underlying navigational graph (as in Fig.~\ref{Fig:graph_split}) with additional visual information (as in Fig.~\ref{fig:intro}).
In order to test whether the agent model could overfit to these navigational graphs (and thus be biased towards training environments), we follow the experiments in \cite{hu2019you} to train the agent without visual information,
masking out the ResNet features which are used as the visual input with zero vectors.
When the agent could only make the decision based on the instruction and the navigational graph, 
our baseline model gives the success rate of $38.5\%$ and $41.0\%$ on validation seen and unseen in this setting, which is consistent with the finding in \cite{hu2019you}.
Besides showing the relatively good performance on unseen split without visual contents (similar to \cite{thomason2018shifting} and \cite{hu2019you}), the model generates the low performance gap between seen and unseen environments ($2.5\%$ compared to the $>10\%$ gap in usual).
Hence, we claim that the underlying graph is not a dominant reason for the environment bias.

\subsection{The Effect of Visual Environments}
To show how the visual information plays its role in environment bias, we analyze agent's performance on unseen environments and in different spatial regions of the training environments.
In order to give a detailed characterization of the effect of environments, we will reveal three different levels of spatial localities which are related to the agent's performance:
\begin{itemize}
\item \textbf{Path-level Locality:} Agents are better at paths which intersect with the training paths.
\item \textbf{Region-level Locality:} Agents are better in regions that are closer to the training data.
\item \textbf{Environment-level Locality:} Agents perform better in training environments than in unseen environments.
\end{itemize}
The existence of these spatial locality inspires us to find the direct cause of the problem in Sec.~\ref{sec-3-2-envbias}.
However, the original split of data is not fine-grained enough to separately reveal these spatial localities.
As we visualize the data from one environment of the Room-to-Room dataset in the left part of Fig.~\ref{Fig:graph_split}, where the vertices are viewpoints with visual information and edges are valid connections between viewpoints, 
nearly all viewpoints in val-seen paths (vertices connected to red lines) are also included in training paths (vertices marked by dark-black).
We thus cannot categorize the path-level and region-level localities.
To bypass this, we propose a novel re-splitting method to create our diagnosis data splits.

\paragraph{Structural Data Re-splitting}
We employ two kinds of structural data splitting methods based on the horizontal or vertical coordinates, denoted as `X-split' and `Z-split' respectively.
The `Z-split' separates different floors of houses and `X-split' creates separate areas.
When applying to training environments in R2R dataset, we use one side of the splitting line (see `X-splitting line' in Fig.~\ref{Fig:graph_split}) as the new training `environment', the other side as the path-unseen `environment'.
The original training and val-seen data are re-split accordingly, while the val-unseen data and environments keep untouched.
As shown in the right part of Fig.~\ref{Fig:graph_split}, our re-splitting method creates three new data splits: training split, val-path-seen split (i.e., data intersecting with training paths), and val-path-unseen split  
(i.e., data not intersecting with training paths but located in the same original environments). 
The original val-unseen set is denoted as `env-unseen' split.

\begin{table}[t]
\small
\begin{center}
    \begin{tabular}{lccccc}
        & 
        \multirow{2}{*}{\bf Splitting}
        & \multirow{2}{*}{\bf Train} & \multicolumn{3}{c}{\bf\small Validation}\\
    \cmidrule(lr){4-6}
        &       & & \bf \small P-seen  & \bf  \small P-unseen  & \small \bf  E-unseen  \\
    \midrule
    \multirow{3}{*}{\bf Env} 
        & R2R   & 61           & 56                & 0               & 11 \\
        & X-split  & 61           & 57                & 16               & 11 \\
        & Z-split  & 61           & 56                & 29               & 11 \\
    \midrule
    \multirow{3}{*}{\bf Data}
        & R2R   & 14,025        & 1,020              &  0              & 2,349 \\
        & X-split  & 11,631        & 1,230              & 1,098             & 2,349 \\
        & Z-split  & 10,894        & 867              & 2,324             & 2,349 \\
    \midrule
    \multirow{3}{*}{\bf SR}
        & R2R    & 88.3          & 56.1              & --               & 47.5 \\
        & X-split   & 87.3  & 58.9              & 52.6             & 46.7 \\
        & Z-split   & 94.7  & 62.5              & 47.8             & 42.4 \\
    \end{tabular}
\end{center}
\caption{
The statistics and results of original R2R splits and our new splits, showing the path-level and environment-level localities. 
Numbers of environments and data are denoted as `Env' and `Data'; success rate is denoted as `SR'; val-path-seen, val-path-unseen and val-env-unseen are denoted as `P-seen', `P-unseen' and `E-unseen'.
}
\label{x-split}
\end{table}

\paragraph{Existence of Path-level and Environment-level Localities}
For both splitting methods, we train our baseline model on the newly-split training set and evaluate it on the three validation sets.
As shown in Table~\ref{x-split}, the agent performs better on val path-seen than val path-unseen, 
which suggests that a \textbf{path-level locality} exists in current VLN agent models.
Meanwhile, the results on val path-unseen are then higher than val env-unseen, indicating the existence of \textbf{environment-level locality} which is independent of the path-level locality.

\paragraph{Existence of Region-level Locality}
To demonstrate region-level locality,
we study how the success rate changes with respect to distances between training and validation environment regions,
similar to the analysis of language `distance' in Sec.~\ref{sec-4-1-lang}.
We calculate the point-by-point shortest paths using the Dijkstra's algorithm, where the shortest distances between viewpoints $v$ and $v'$ are denoted as the graph distance $\mathrm{dis}_\textsc{graph}(v, v')$.
We first define the viewpoint distance $\mathrm{dis}_\textsc{viewpoint}$ from a viewpoint $v$ to the training data $\mathbb{T}$ as the minimal graph distance from $v$ to a viewpoint $v'$ in training data;
Then the path distance $\mathrm{dis}_\textsc{path}$ from a validating data $x$ to the whole training data $\mathbb{T}$ is defined as the maximal viewpoint distance of the viewpoints in the path of $x$:
\begin{align}
\mathrm{dis}_\textsc{path}(x, \mathbb{T}) &= \max_{v \in \mathrm{path}(x)} \mathrm{dis}_\textsc{viewpoint} (v, \mathbb{T}) \\
\mathrm{dis}_\textsc{viewpoint} (v, \mathbb{T}) &= \min_{\substack{v' \in \mathrm{path}(t) \\ \forall  t \in \mathbb{T}}}
\mathrm{dis}_\textsc{graph}\left(v, v'\right)
\end{align}
We compute path distances between paths in the path-unseen validation set and training environments of our re-splitting data.
As shown in Table~\ref{region_locality}, the success rate declines as the path moves further from the training environment in both re-splitting methods (i.e., `X-split' and `Z-split').
Therefore, the closer the path to the training data, the better the agent performs, which reveals the existence of \textbf{region-level locality}.

\begin{table}[t] \small
\begin{center}
    \begin{tabular}{llcccc}
    \multirow{2}{*}{\bf X-split} & \bf PD (meters)
        & 5-13 & 14-16 & 17-21 & 22-57 \\
    \cmidrule(lr){2-6}
        & \bf SR (\%) & 56.3 & 56.2 & 50.5 & 43.6  \\
    \midrule
    \multirow{2}{*}{\bf Z-split} & \bf PD (meters)
        & 5-10 & 11-13 & 14-17 & 18-52 \\
    \cmidrule(lr){2-6}
        & \bf SR (\%) & 58.5 & 47.9 & 42.9 & 44.1  \\
    \end{tabular}
\end{center}
\caption{The success rate declines as the path moves further from training regions. Path distance is denoted as `PD'.}
\label{region_locality}
\end{table}
\section{Why: What Inside the Environments Contributes to the Bias?} 
\label{sec-5-why}
In Sec.~\ref{sec-4-where}, we locate the cause of the performance gap in visual environments by excluding other potential reasons and categorizing the spatial localities. 
However, there are still multiple possible aspects inside the environment which could lead to these spatial localities, e.g., the objects layout and room connections. 
The agent could be biased towards the training environments by over-memorizing these environment-specific characteristics.
In this section, we pinpoint the aspect that directly contributes to the bias to be the low-level visual information carried by the ResNet features.
We first show an experiment where the gap between seen and unseen environments is effectively decreased by minimal model modifications.
We then clarify our conclusions based on the findings. 

\subsection{An Investigation Experiment: ImageNet Labels as Visual Features }
\label{sec-5-1}
Suspecting that the over-fitting is caused by low-level ResNet 2048-features which the agent over-learns,
we hope to find a replacement that contains minimal low-level information while preserving distinguishable visual contents.
The most straightforward replacement is that instead of using mean-pooled features, we inject the frozen 1000-way classifying layer from ResNet pre-training, and use the probabilities of ImageNet labels as visual features.
Shown as `ImageNet' in Table~\ref{results}, 
the probability distribution reduces the gaps between seen and unseen in all three datasets, and almost closes the gaps of R2R and R4R.
These results further reveal the low-level ResNet features of image views as the reason for environment bias.
Combining with the findings of spatial localities, we suggest that environments (i.e., houses) and regions (i.e., rooms) usually have their own `styles'.
Thus the same semantic label (captured by ImageNet-1000 features) may have different visual appearances (captured by ResNet features) in different environments or regions.
As a result, ImageNet-1000 features, in spite of being noisy, are not distracted by low-level visual appearance and thus could generalize well to unseen environments, while ResNet features could not.

Although this ImageNet-1000-feature replacement could decrease the performance gap, it has a disagreement with the VLN domain that the validation unseen result of R4R is slightly worse than baseline (and not much better in R2R and CVDN cases).
Hence it motivates us to find better semantic representations of environmental features that can both close the seen-unseen gap while also achieving state-of-the-art on unseen results (which we will discuss next).

\begin{table}[t] \small
\newcommand{\tabincell}[2]{\begin{tabular}{@{}#1@{}}#2\end{tabular}}
\begin{center}
\begin{tabular}{clccc}
\multirow{2}{*}{\bf Task}  &\multirow{2}{*}{\bf Feature} & \multicolumn{3}{c}{\bf Result} \\
\cmidrule(lr){3-5}
& & \multicolumn{1}{c}{\bf Val Seen}  &\multicolumn{1}{c}{\bf Val Unseen} & 
\multicolumn{1}{c}{\bf Gap $\lvert \Delta\rvert$ }
\\ 
\midrule
\multirow{5}{*}{R2R}  
                & ResNet         & 56.1  & 47.5   & 8.6   \\
                & ImageNet         & 47.1  & 48.2   & 1.1   \\
                & Detection     & 55.9  & 50.0   & 5.9   \\
                & GT-Seg   & 55.6  & 56.2   & 0.6   \\
                & Learned-Seg         & 52.6  & 53.3   & 0.7   \\
\midrule                                            
\multirow{5}{*}{R4R}        
                & ResNet          & 54.6  & 30.7  & 23.9  \\
                & ImageNet         & 28.7  & 28.9  & 0.2   \\
                & Detection    & 48.8  & 32.0  & 16.8   \\
                & GT-Seg   & 47.6  & 35.9  & 11.7  \\
                & Learned-Seg          & 38.0  & 34.3  & 3.7   \\
\midrule                                            
\multirow{5}{*}{CVDN}                 
                & ResNet         & 6.60  &  3.05 & 3.55  \\
                & ImageNet        &   5.70    &   3.11    &  2.59     \\
                & Detection        & 6.55  &  3.94 & 2.61  \\
                & GT-Seg     & 6.46  &  4.33 & 2.13  \\
                & Learned-Seg        & 5.82  &  4.41 & 1.41  \\
\end{tabular}
\end{center}
\caption{Results showing that our semantic feature representations, i.e., ImageNet, Detection, ground-truth and learned semantic segmentation (denoted as `GT-Seg' and `Learned-Seg'), effectively reduce the performance gaps in all three datasets.
}
\label{results}
\end{table}

\section{How: Methodology to Fix Environment Bias}
\label{sec-6}
In the previous section (Sec.~\ref{sec-5-why}), we found that the environment bias is related to the low-level visual information (i.e., $2048$-dim ResNet features),
and we want to build our agent on the features which are correlated to the VLN environmental semantics following the observation in Sec.~\ref{sec-5-1}.
In this section, several advanced semantic feature replacements are explored.
As shown in Table~\ref{results}, these features could effectively reduce the performance gaps between seen and unseen environments and improve the unseen results compared to our strong baselines, without any changes in training procedure or hyperparameters.
The effectiveness of these semantic representations supports our explanation of the environment bias in Sec.~\ref{sec-5-why} and also suggests that future work in VLN tasks should think about such generalization issues.

\subsection{Baseline}
In our baseline model (described in Sec.~\ref{sec:VLN}), following the previous works we use the standard ResNet features as the representation of environments~\cite{anderson2018vision,jain2019stay,thomason2019vision}.
These features come from the mean-pooled layer after the final convolutional layer of ResNet-152~\cite{he2016deep} pre-trained on ImageNet~\cite{russakovsky2015imagenet}.
As shown in `Baseline' rows of Table~\ref{results}, in this setting, val-seen results are significantly higher than val-unseen results in all three datasets.
The `environment bias' phenomenon, which we described in Sec.~\ref{sec-3-2-envbias}, is observed in our baseline model, leading us to the following discussions of semantic features.

\subsection{Detected Object Areas} 
During navigation, the objects in the environments are crucial since their matchings with instruction often indicate the locations that can guide the agent,
thus object detection quality of the environments can provide relevant semantic information.
In our work, we utilize the detection information generated by Faster R-CNN~\cite{ren2015faster} to create the feature representations.
Compared to ImageNet-1000 features (Sec.~\ref{sec-5-1}), these detection features include more environmental information since the viewing images in VLN usually contain multiple objects.
Instead of directly using classification probabilities
or the label embeddings (in \cite{hu2019you}), 
we design our features $\mathrm{f}_\textsc{\,detect}$ of each image view as the collection of the area summations weighted by detection confidence of each detected object:
\begin{align}
\mathrm{f}_\textsc{\,detect}  \mbox{=} [a_{c_1}, a_{c_2}, \ldots, a_{c_n}]; \\
a_{c_i} \mbox{=} \sum_{\mathrm{obj} \text{ is } c_i}  \mathrm{Area}(\mathrm{obj}) \cdot \mathrm{Conf}(\mathrm{obj})
\end{align}
where $a_{c_i}$ is feature of the object label $c_i$.
The area and confidence are denoted as $\mathrm{Area}(*)$ and $\mathrm{Conf}(*)$.
For implementation details, we use the Faster R-CNN~\cite{ren2015faster} trained on Visual Genome~\cite{krishna2017visual} provided in Bottom-Up Attention~\cite{anderson2018bottom}.
To eliminate labels irrelevant to the VLN task, we calculate the total areas of each detection object among all environments and pick the labels that take up a relatively large proportion of the environments, creating features of dimension $152$.
The results are denoted as `Detection' in Table~\ref{results}. 
The performance gaps are diminished with these detection features compared to baselines in all three datasets, indicating that lifting the features to a higher semantic level has a positive effect on alleviating the environment bias.
Meanwhile, the improvement of unseen validation results suggests better effectiveness in the VLN task than the ImageNet labels.

\subsection{Semantic Segmentation}
\label{sec-6-3-semantic-view}
Although the detection features can provide adequate semantic information for the agent to achieve comparable results as the baseline model, they do not fully utilize the visual information where the content left over from detection may contain useful knowledge for navigation.
A better semantic representation is the semantic segmentation, which segments each view image on the pixel level and gives the label to each segment region.
Matterport3D~\cite{chang2017matterport3d} dataset provides the labeled semantic segmentation information of every scene and we take the rendered images from \cite{tan2019learning}.
An example of comparison between RGB images and semantic segmentation is available in the Appendix.
Since the semantic segmentation images are noisy and blurry in boundaries, following the design of detection features, the semantic features are designed as the areas of semantic classes in each image view, with dimension of $42$. 
We first assume that the semantic segmentation is provided as additional environmental information and the results of the model using the ground-truth semantic segmentation areas are shown in the `GT-Seg' rows in Table~\ref{results}.
We next study the situation where the semantic information is not available in testing environments thus the information needs to be learned from training environments.
A separate multi-layer perceptron is trained to predict the areas of these semantic classes from 2048-dim ResNet features.
This MLP is trained on 51 (out of 61) training environments and its hyper-parameters are validated on the remaining training 10 environments (and more details are available in the Appendix).
The results of the agent model with these predicted semantics as features are shown in `Learned-Seg'
As shown in Table~\ref{results}, both `GT-Seg' and `Learned-Seg' semantic representations bring the performance of seen and unseen even closer compared to the previous detection semantics.
The highest validation unseen success rates among our proposed representations are also produced by these methods: `Learned-Seg' semantics for CVDN and `GT-Seg' semantics for R2R and R4R. 
Overall, among all explored visual representations, the semantic segmentation features are most effective in reducing the environment bias.\footnote{To further prove the effectiveness of our semantic features in alleviating the environment bias, we test our ‘Learned-Seg’ features on two other methods from previous works (SMNA and Regretful). See details in Appendix.}

\section{Conclusion}
In this paper, we focus on studying the performance gap between seen and unseen environments widely observed in vision-and-language navigation (VLN) tasks, trying to find where and why this environment bias exists and provide possible initial solutions.
By designing the diagnosis experiments of environment re-splitting and feature replacement, we locate the environment bias to be in the low-level visual appearance; and we discuss semantic features that decrease the performance gaps in three VLN datasets and achieve state-of-the-art results.

\section*{Acknowledgments}
We thank P. Hase, H. Guo, L. Yu, S. Zhang, and X. Zhou for their helpful suggestions. This work was supported by ARO-YIP Award W911NF-18-1-0336, ONR Grant N00014-18-1-2871, NIH grant 1R01LM013329-01, and a Google Focused Research Award. The views, opinions, and/or findings contained in this article are those of the authors and should not be interpreted as representing the official views or
policies, either expressed or implied, of the funding agency.

\appendix
\section{Example of RGB Images and Semantic Segmentation}

\begin{figure}[t]
\centering
\includegraphics[width=0.45\textwidth]{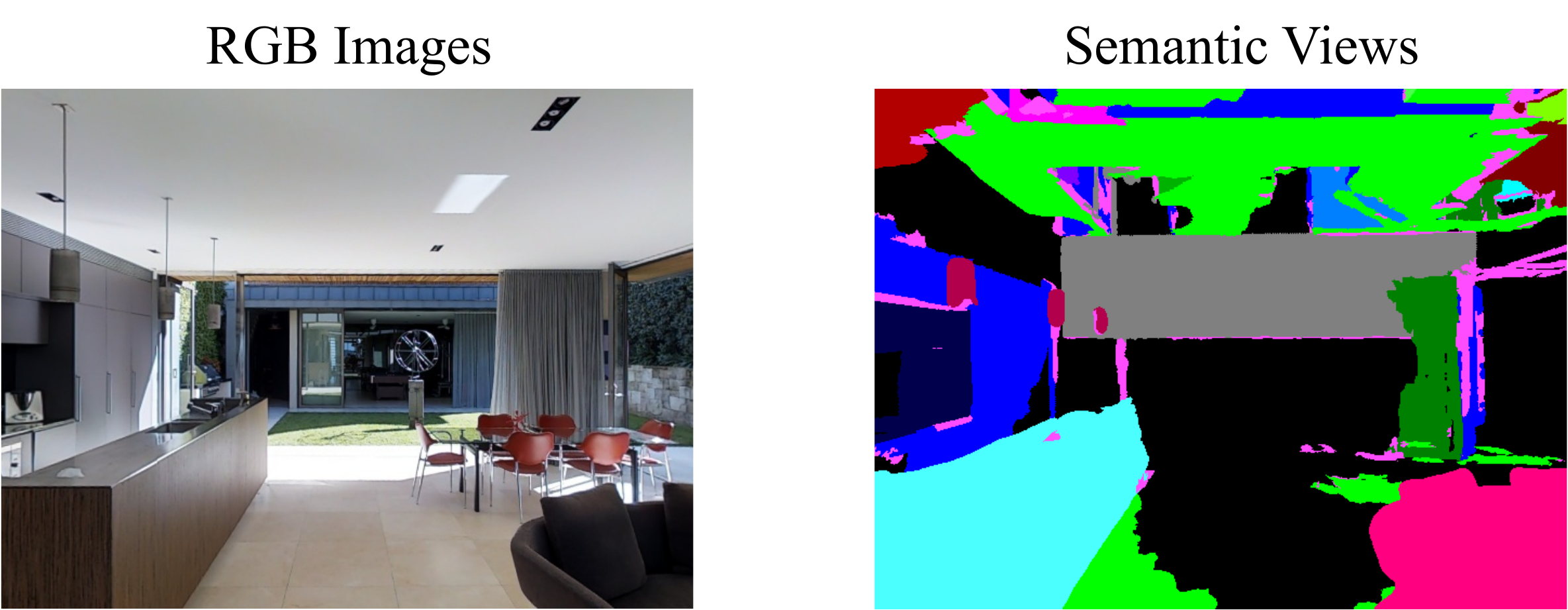} 
\caption{Comparisons between RGB images and their semantic views.}
\label{Fig:semantic_view}
\end{figure}

In Fig.~\ref{Fig:semantic_view}, we show a rendered semantic view from \cite{tan2019learning} and its original RGB image.
Different colors indicate the label of segmentation areas.
$42$ semantic labels ($40$ semantic classes and $2$ structural classes) are considered in the Matterport3D dataset \cite{chang2017matterport3d}.

\section{Details of `Learned-Seg' Semantic Training}
We use a multi-layer perceptron (MLP) to generate the `Learned-Seg' semantic features.
The multi-layer perceptron includes three fully-connected layers with $\mathrm{ReLU}$ activation on the outputs of the first two layers.
The input of this MLP is the $2048$-dim ResNet feature $f$ of each image view. The hidden sizes of the first two layers are $512$ and $256$. The final layer will output the $42$-dim semantic feature $y$ that represents the areas of each semantic class.
After the linear layers, we use the sigmoid function $\sigma$ to convert the output to the ratio of areas.
\begin{align}
    x_1 &= \mathrm{ReLU}( A_1 f + b_1) \\
    x_2 &= \mathrm{ReLU}( A_2 x_1 + b_2) \\
    y &= \sigma (A_3 x_2 + b_3)
\end{align}

The model is trained with ground-truth semantic areas (normalized to $[0,1]$) of the views in $51$ environments out of total $61$ VLN training environments, and is tuned on the remaining $10$ environments.
We minimize the binary cross-entropy loss between the ground-truth areas $\{y^*_i\}$ and the predicted areas $\{y_i\}$, where $i$ indicate the $i$-th semantic class.
\begin{align}
    \mathcal{L} = -\sum_i \left( y^*_i \log y_i + \left(1 - y^*_i\right) \log \left(1 - y_i\right) \right)
\end{align}
Dropout layers with a probability of $0.3$ are added between fully-connected layers while training.

After the model is fitted, we freeze the weight and use it to predict the `Learned-Seg' semantic features of all seen and unseen environments (i.e., environments of all training, val-seen, and val-unseen data).
These extracted features are further used in training the neural navigational agent model.

\begin{table}[t] \small
\begin{center}
    \begin{tabular}{llccc}
        \multirow{2}{*}{\bf Models} & \multirow{2}{*}{\bf Features} & \multicolumn{3}{c}{\bf Result} \\
        \cmidrule(lr){3-5}
        & & \multicolumn{1}{c}{\bf Val Seen}  &\multicolumn{1}{c}{\bf Val Unseen} & \multicolumn{1}{c}{\bf Gap $\lvert \Delta\rvert$ } \\
    \midrule
    \multirow{2}{*}{SMNA} 
        & ResNet & 63.9 & 38.0 & 25.9 \\
    \cmidrule(lr){2-5}
        & Learned-Seg & 55.0 & 40.6 & 14.4  \\
    \midrule
    \multirow{2}{*}{Regretful} 
        & ResNet & 63.8 & 42.7 & 21.1 \\
    \cmidrule(lr){2-5}
        & Learned-Seg & 55.2 & 47.3 & 7.9  \\
    \end{tabular}
\end{center}
\caption{When being used in two other agent models, our learned semantic segmentation features still show the effectiveness in alleviating the environment bias.}
\label{other-baseline}
\end{table}

\section{Other Navigational Models with `Learned-Seg' Features}
To further prove the effectiveness of our semantic features in alleviating the environment bias, we test our `Learned-Seg' features (described in Sec.~\ref{sec-6-3-semantic-view}) on two other baselines from previous works, i.e., SMNA~\cite{ma2019self} and Regretful~\cite{ma2019regretful}.\footnote{We use their official code implementation in \href{https://github.com/chihyaoma/selfmonitoring-agent}{https://github.com/chihyaoma/selfmonitoring-agent} and \href{https://github.com/chihyaoma/regretful-agent}{https://github.com/chihyaoma/regretful-agent}}
As shown in Table~\ref{other-baseline}, when being used in different neural agent models, our semantic features can also effectively reduce the performance gaps as well as bring extra improvement on val unseen results.

\bibliographystyle{named}
\bibliography{short_cite}

\begin{thebibliography}{}

\bibitem[\protect\citeauthoryear{Anderson \bgroup \em et al.\egroup
  }{2018a}]{anderson2018bottom}
P.~Anderson, X.~He, et~al.
\newblock Bottom-up and top-down attention for image captioning and visual
  question answering.
\newblock In {\em CVPR}, 2018.

\bibitem[\protect\citeauthoryear{Anderson \bgroup \em et al.\egroup
  }{2018b}]{anderson2018vision}
P.~Anderson, Q.~Wu, et~al.
\newblock Vision-and-language navigation: Interpreting visually-grounded
  navigation instructions in real environments.
\newblock In {\em CVPR}, 2018.

\bibitem[\protect\citeauthoryear{Anderson \bgroup \em et al.\egroup
  }{2019}]{anderson2019chasing}
P.~Anderson, A.~Shrivastava, et~al.
\newblock Chasing ghosts: Instruction following as bayesian state tracking.
\newblock In {\em NeurIPS}, 2019.

\bibitem[\protect\citeauthoryear{Bahdanau \bgroup \em et al.\egroup
  }{2015}]{bahdanau2014neural}
D.~Bahdanau, K.~Cho, and Y.~Bengio.
\newblock Neural machine translation by jointly learning to align and
  translate.
\newblock In {\em ICLR}, 2015.

\bibitem[\protect\citeauthoryear{Blanchard \bgroup \em et al.\egroup
  }{2011}]{blanchard2011generalizing}
G.~Blanchard, G.~Lee, and C.~Scott.
\newblock Generalizing from several related classification tasks to a new
  unlabeled sample.
\newblock In {\em NeurIPS}, 2011.

\bibitem[\protect\citeauthoryear{Blukis \bgroup \em et al.\egroup
  }{2018}]{blukis2018mapping}
V.~Blukis, D.~Misra, et~al.
\newblock Mapping navigation instructions to continuous control actions with
  position-visitation prediction.
\newblock In {\em CoRL}, 2018.

\bibitem[\protect\citeauthoryear{Carlucci \bgroup \em et al.\egroup
  }{2019}]{carlucci2019domain}
F.~M. Carlucci, A.~D'Innocente, et~al.
\newblock Domain generalization by solving jigsaw puzzles.
\newblock In {\em CVPR}, 2019.

\bibitem[\protect\citeauthoryear{Chang \bgroup \em et al.\egroup
  }{2017}]{chang2017matterport3d}
A.~Chang, A.~Dai, et~al.
\newblock Matterport3d: Learning from rgb-d data in indoor environments.
\newblock In {\em 3DV}, 2017.

\bibitem[\protect\citeauthoryear{Chen \bgroup \em et al.\egroup
  }{2019a}]{chen2019joint}
C.~Chen, Z.~Chen, et~al.
\newblock Joint domain alignment and discriminative feature learning for
  unsupervised deep domain adaptation.
\newblock In {\em AAAI}, 2019.

\bibitem[\protect\citeauthoryear{Chen \bgroup \em et al.\egroup
  }{2019b}]{chen2019touchdown}
H.~Chen, A.~Suhr, et~al.
\newblock Touchdown: Natural language navigation and spatial reasoning in
  visual street environments.
\newblock In {\em CVPR}, 2019.

\bibitem[\protect\citeauthoryear{Das \bgroup \em et al.\egroup
  }{2018}]{das2018embodied}
A.~Das, S.~Datta, et~al.
\newblock Embodied question answering.
\newblock In {\em CVPR}, 2018.

\bibitem[\protect\citeauthoryear{Fried \bgroup \em et al.\egroup
  }{2018}]{fried2018speaker}
D.~Fried, R.~Hu, et~al.
\newblock Speaker-follower models for vision-and-language navigation.
\newblock In {\em NeurIPS}, 2018.

\bibitem[\protect\citeauthoryear{Goodfellow \bgroup \em et al.\egroup
  }{2014}]{goodfellow2014generative}
I.~Goodfellow, J.~Pouget-Abadie, et~al.
\newblock Generative adversarial nets.
\newblock In {\em NeurIPS}, 2014.

\bibitem[\protect\citeauthoryear{He \bgroup \em et al.\egroup
  }{2016}]{he2016deep}
K.~He, X.~Zhang, et~al.
\newblock Deep residual learning for image recognition.
\newblock In {\em CVPR}, 2016.

\bibitem[\protect\citeauthoryear{Hu \bgroup \em et al.\egroup
  }{2019}]{hu2019you}
R.~Hu, D.~Fried, et~al.
\newblock Are you looking? grounding to multiple modalities in
  vision-and-language navigation.
\newblock In {\em ACL}, 2019.

\bibitem[\protect\citeauthoryear{Huang \bgroup \em et al.\egroup
  }{2019}]{huang2019transferable}
H.~Huang, V.~Jain, et~al.
\newblock Transferable representation learning in vision-and-language
  navigation.
\newblock In {\em ICCV}, 2019.

\bibitem[\protect\citeauthoryear{Jain \bgroup \em et al.\egroup
  }{2019}]{jain2019stay}
V.~Jain, G.~Magalhaes, et~al.
\newblock Stay on the path: Instruction fidelity in vision-and-language
  navigation.
\newblock In {\em ACL}, 2019.

\bibitem[\protect\citeauthoryear{Ke \bgroup \em et al.\egroup
  }{2019}]{ke2019tactical}
L.~Ke, X.~Li, et~al.
\newblock Tactical rewind: Self-correction via backtracking in
  vision-and-language navigation.
\newblock In {\em CVPR}, 2019.

\bibitem[\protect\citeauthoryear{Krishna \bgroup \em et al.\egroup
  }{2017}]{krishna2017visual}
R.~Krishna, Y.~Zhu, et~al.
\newblock Visual genome: Connecting language and vision using crowdsourced
  dense image annotations.
\newblock {\em IJCV}, 123(1):32--73, 2017.

\bibitem[\protect\citeauthoryear{Lin}{2004}]{lin2004rouge}
C.~Lin.
\newblock Rouge: A package for automatic evaluation of summaries.
\newblock In {\em Text summarization branches out}, pages 74--81, 2004.

\bibitem[\protect\citeauthoryear{Long \bgroup \em et al.\egroup
  }{2018}]{long2018conditional}
M.~Long, Z.~Cao, et~al.
\newblock Conditional adversarial domain adaptation.
\newblock In {\em NeurIPS}, 2018.

\bibitem[\protect\citeauthoryear{Ma \bgroup \em et al.\egroup
  }{2019a}]{ma2019self}
C.~Ma, J.~Lu, et~al.
\newblock Self-monitoring navigation agent via auxiliary progress estimation.
\newblock In {\em ICLR}, 2019.

\bibitem[\protect\citeauthoryear{Ma \bgroup \em et al.\egroup
  }{2019b}]{ma2019regretful}
C.~Ma, Z.~Wu, et~al.
\newblock The regretful agent: Heuristic-aided navigation through progress
  estimation.
\newblock In {\em CVPR}, 2019.

\bibitem[\protect\citeauthoryear{MacMahon \bgroup \em et al.\egroup
  }{2006}]{macmahon2006walk}
M.~MacMahon, B.~Stankiewicz, and B.~Kuipers.
\newblock Walk the talk: connecting language, knowledge, and action in route
  instructions.
\newblock In {\em AAAI}, 2006.

\bibitem[\protect\citeauthoryear{Mirowski \bgroup \em et al.\egroup
  }{2018}]{mirowski2018learning}
P.~Mirowski, M.~Grimes, et~al.
\newblock Learning to navigate in cities without a map.
\newblock In {\em NeurIPS}, 2018.

\bibitem[\protect\citeauthoryear{Muandet \bgroup \em et al.\egroup
  }{2013}]{muandet2013domain}
K.~Muandet, D.~Balduzzi, and B.~Sch{\"o}lkopf.
\newblock Domain generalization via invariant feature representation.
\newblock In {\em ICML}, 2013.

\bibitem[\protect\citeauthoryear{Papineni \bgroup \em et al.\egroup
  }{2002}]{papineni2002bleu}
K.~Papineni, S.~Roukos, et~al.
\newblock Bleu: a method for automatic evaluation of machine translation.
\newblock In {\em ACL}, 2002.

\bibitem[\protect\citeauthoryear{Ren \bgroup \em et al.\egroup
  }{2015}]{ren2015faster}
S.~Ren, K.~He, et~al.
\newblock Faster r-cnn: Towards real-time object detection with region proposal
  networks.
\newblock In {\em NeurIPS}, 2015.

\bibitem[\protect\citeauthoryear{Rozantsev \bgroup \em et al.\egroup
  }{2018}]{rozantsev2018residual}
A.~Rozantsev, M.~Salzmann, and P.~Fua.
\newblock Residual parameter transfer for deep domain adaptation.
\newblock In {\em CVPR}, 2018.

\bibitem[\protect\citeauthoryear{Russakovsky \bgroup \em et al.\egroup
  }{2015}]{russakovsky2015imagenet}
O.~Russakovsky, J.~Deng, et~al.
\newblock Imagenet large scale visual recognition challenge.
\newblock {\em IJCV}, 115(3):211--252, 2015.

\bibitem[\protect\citeauthoryear{Tan \bgroup \em et al.\egroup
  }{2019}]{tan2019learning}
H.~Tan, L.~Yu, and M.~Bansal.
\newblock Learning to navigate unseen environments: Back translation with
  environmental dropout.
\newblock In {\em NACCL}, 2019.

\bibitem[\protect\citeauthoryear{Thomason \bgroup \em et al.\egroup
  }{2019a}]{thomason2018shifting}
J.~Thomason, D.~Gordan, and Y.~Bisk.
\newblock Shifting the baseline: Single modality performance on visual
  navigation \& qa.
\newblock In {\em NACCL}, 2019.

\bibitem[\protect\citeauthoryear{Thomason \bgroup \em et al.\egroup
  }{2019b}]{thomason2019vision}
J.~Thomason, M.~Murray, et~al.
\newblock Vision-and-dialog navigation.
\newblock In {\em CoRL}, 2019.

\bibitem[\protect\citeauthoryear{Wang \bgroup \em et al.\egroup
  }{2018}]{wang2018look}
X.~Wang, W.~Xiong, et~al.
\newblock Look before you leap: Bridging model-free and model-based
  reinforcement learning for planned-ahead vision-and-language navigation.
\newblock In {\em ECCV}, 2018.

\bibitem[\protect\citeauthoryear{Wang \bgroup \em et al.\egroup
  }{2019}]{wang2019reinforced}
X.~Wang, Q.~Huang, et~al.
\newblock Reinforced cross-modal matching and self-supervised imitation
  learning for vision-language navigation.
\newblock In {\em CVPR}, 2019.

\bibitem[\protect\citeauthoryear{Yu \bgroup \em et al.\egroup
  }{2019}]{yu2019multi}
L.~Yu, X.~Chen, et~al.
\newblock Multi-target embodied question answering.
\newblock In {\em CVPR}, 2019.

\end{thebibliography}

\end{document}